\documentclass{ecai} 

\usepackage{latexsym}
\usepackage{amssymb}
\usepackage{amsmath}
\usepackage{amsthm}
\usepackage{booktabs}
\usepackage{enumitem}
\usepackage{graphicx}
\usepackage{color}
\usepackage{algorithm} %
\usepackage{algpseudocode} %
\usepackage{multirow} %
\usepackage{makecell} %
\usepackage{algpseudocode} %
\usepackage{array}

\newcommand{\BibTeX}{B\kern-.05em{\sc i\kern-.025em b}\kern-.08em\TeX}

\begin{document}


\begin{frontmatter}


\paperid{1027} 


\title{A New One-Shot Federated Learning Framework for Medical Imaging Classification with Feature-Guided Rectified Flow and Knowledge Distillation}



\author[A]{\fnms{Yufei}~\snm{Ma}}
\author[A]{\fnms{Hanwen}~\snm{Zhang}}
\author[A]{\fnms{Qiya}~\snm{Yang}}
\author[A]{\fnms{Guibo}~\snm{Luo}\footnote[*]{Corresponding Author. Email: \{luogb, zhuys\}@pku.edu.cn.}}
\author[A]{\fnms{Yuesheng}~\snm{Zhu}\footnotemark[*]}

\address[A]{Guangdong Provincial Key Laboratory of Ultra High Definition Immersive Media Technology, Shenzhen Graduate School, Peking University}


\begin{abstract}
In multi-center scenarios, One-Shot Federated Learning (OSFL) has attracted increasing attention due to its low communication overhead, requiring only a single round of transmission. However, existing generative model-based OSFL methods suffer from low training efficiency and potential privacy leakage in the healthcare domain. Additionally, achieving convergence within a single round of model aggregation is challenging under non-Independent and Identically Distributed (non-IID) data. To address these challenges, in this paper a modified OSFL framework is proposed, in which a new Feature-Guided Rectified Flow Model (FG-RF) and Dual-Layer Knowledge Distillation (DLKD) aggregation method are developed. FG-RF on the client side accelerates generative modeling in medical imaging scenarios while preserving privacy by synthesizing feature-level images rather than pixel-level images. To handle non-IID distributions, DLKD enables the global student model to simultaneously mimic the output logits and align the intermediate-layer features of client-side teacher models during aggregation. Experimental results on three non-IID medical imaging datasets show that our new framework and method outperform multi-round federated learning approaches, achieving up to 21.73\% improvement, and exceed the baseline FedISCA by an average of 21.75\%. Furthermore, our experiments demonstrate that feature-level synthetic images significantly reduce privacy leakage risks compared to pixel-level synthetic images. The code is available at \url{https://github.com/LMIAPC/one-shot-fl-medical}.
\end{abstract}

\end{frontmatter}


\section{Introduction}





Medical imaging research is often limited by data scarcity and privacy risks \citep{xie2025meddiff, zhou2025federated}. Typically, the amount of training data available in a single hospital or institution is significantly insufficient. However, due to privacy concerns, different medical institutions cannot directly share medical image resources. The development of Federated Learning (FL) \citep{mcmahan2017communication} has addressed these challenges in medical image analysis \citep{luo2022fedsld,liu2024fedfms,zhou2025federated}.

FL is a distributed machine learning framework in which multiple clients independently train local models and share only model updates with a central server for aggregation. This approach effectively protects user data privacy while enabling collaborative training. However, non-Independent and Identically Distributed (non-IID) data across clients remains a major challenge in FL \citep{luo2023influence,luo2022fedsld,zhou2025federated}, significantly affecting aggregation performance. Some studies have attempted to address the non-IID problem, such as FedProx \citep{li2020federated}, SCAFFOLD \citep{karimireddy2020scaffold}, and FedNova \citep{wang2020tackling}, which focus on improving the efficiency and stability of model aggregation by addressing issues like client model drift, gradient inconsistency, and objective misalignment. Nevertheless, these methods rely solely on the raw data available on the client devices, without incorporating any generative models. The reliance on direct model updates and multiple rounds of communication to converge the global model creates significant communication overhead, limiting the scalability and applicability of FL in real-world scenarios. 

One approach to reduce communication overhead is One-Shot Federated Learning (OSFL) \citep{guha2019one}, where clients perform a single round of local training and the global aggregation is completed after just one communication round. To facilitate model aggregation, \citet{guha2019one} and \citet{li2020practical} introduce auxiliary datasets on the server side; however, this strategy is impractical in privacy-sensitive domains such as healthcare \citep{zhang2022dense,chen2024fedbip}. Other methods attempt to transmit intermediate representations of training data \citep{zhou2020distilled,kasturi2020fusion,beitollahi2024parametric,deng2023enhancing} to the server for aggregation, but this approach poses privacy leakage risks.

Nowadays, many studies leverage generative models in OSFL, as they can mitigate these challenges by employing synthetic datasets to update the global model, thereby reducing reliance on auxiliary datasets and minimizing data transfer. Some methods employ generative models such as Variational Autoencoders (VAEs) \citep{heinbaugh2023data} or Generative Adversarial Networks (GANs) \citep{zhang2022dense,kang2023one, kasturi2023osgan,dai2024enhancing}; however, VAEs and GANs often struggle to produce high-quality synthetic data and suffer from unstable training, ultimately affecting the performance of model aggregation. Recently, researchers have proposed to further enhance OSFL performance in natural image analysis \citep{yang2023one,zhang2023federated,yang2024exploring,yang2024feddeo,chen2024fedbip} by integrating pretrained Latent Diffusion Models (LDMs) \citep{rombach2022high}. In contrast, for medical image generation, Denoising Diffusion Probabilistic Models (DDPMs) \citep{ho2020denoising} are generally preferred over fine-tuned LDMs to avoid distribution shift, owing to the significant differences between medical and natural image distributions \citep{zhang2025non}. However, DDPMs suffer from low training efficiency and slow inference speed \citep{liu2022flow}, which limit their practical deployment in FL scenarios. Furthermore, using pixel-level diffusion models still poses a privacy leakage risk, as diffusion models may memorize training data \citep{carlini2023extracting}. Therefore, it is necessary to develop methods that enhance model aggregation performance in OSFL while ensuring data privacy and security.  

\begin{figure}[h]
  \centering
  \begin{minipage}[b]{0.235\textwidth}
    \centering
    \includegraphics[width=\linewidth]{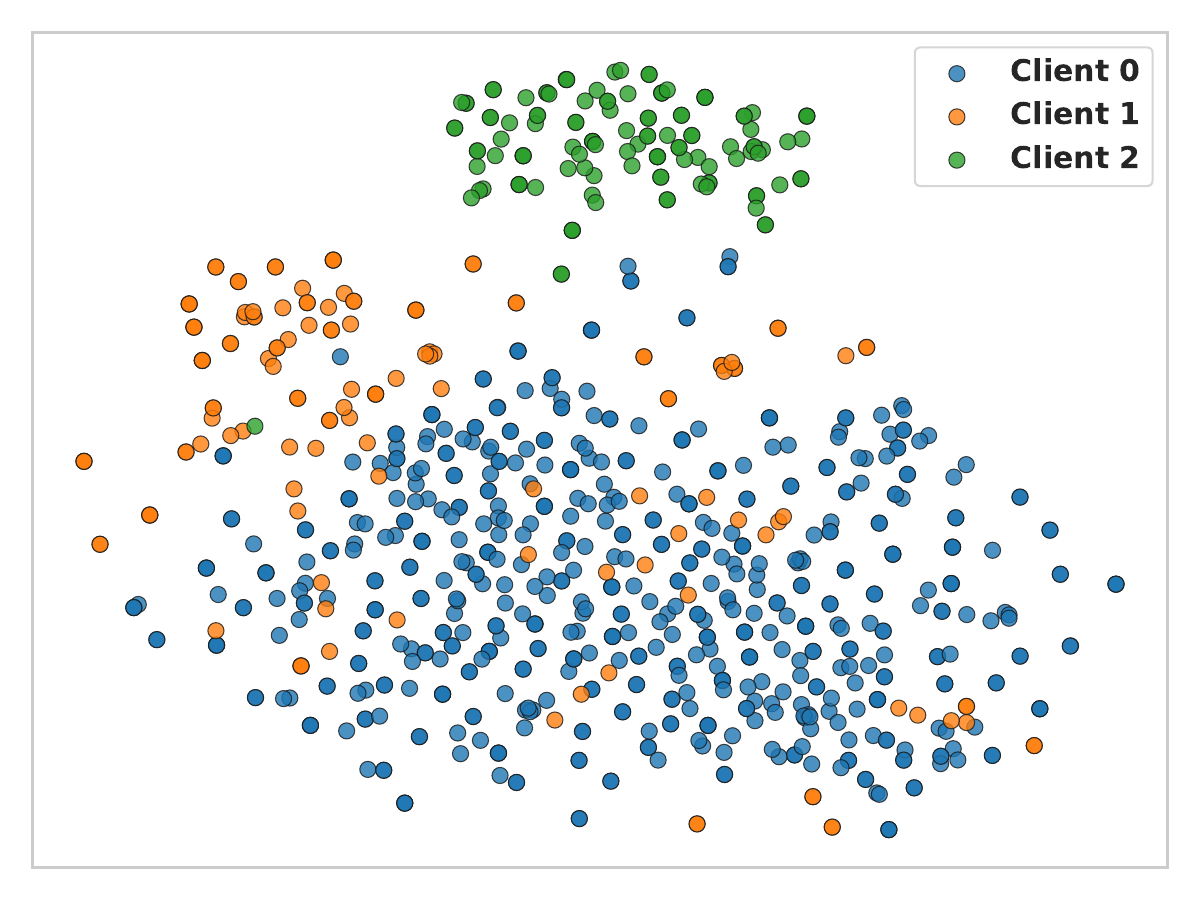}
    \textbf{(a)} Before training
  \end{minipage}
  \hspace{0.05mm}
  \begin{minipage}[b]{0.235\textwidth}
    \centering
    \includegraphics[width=\linewidth]{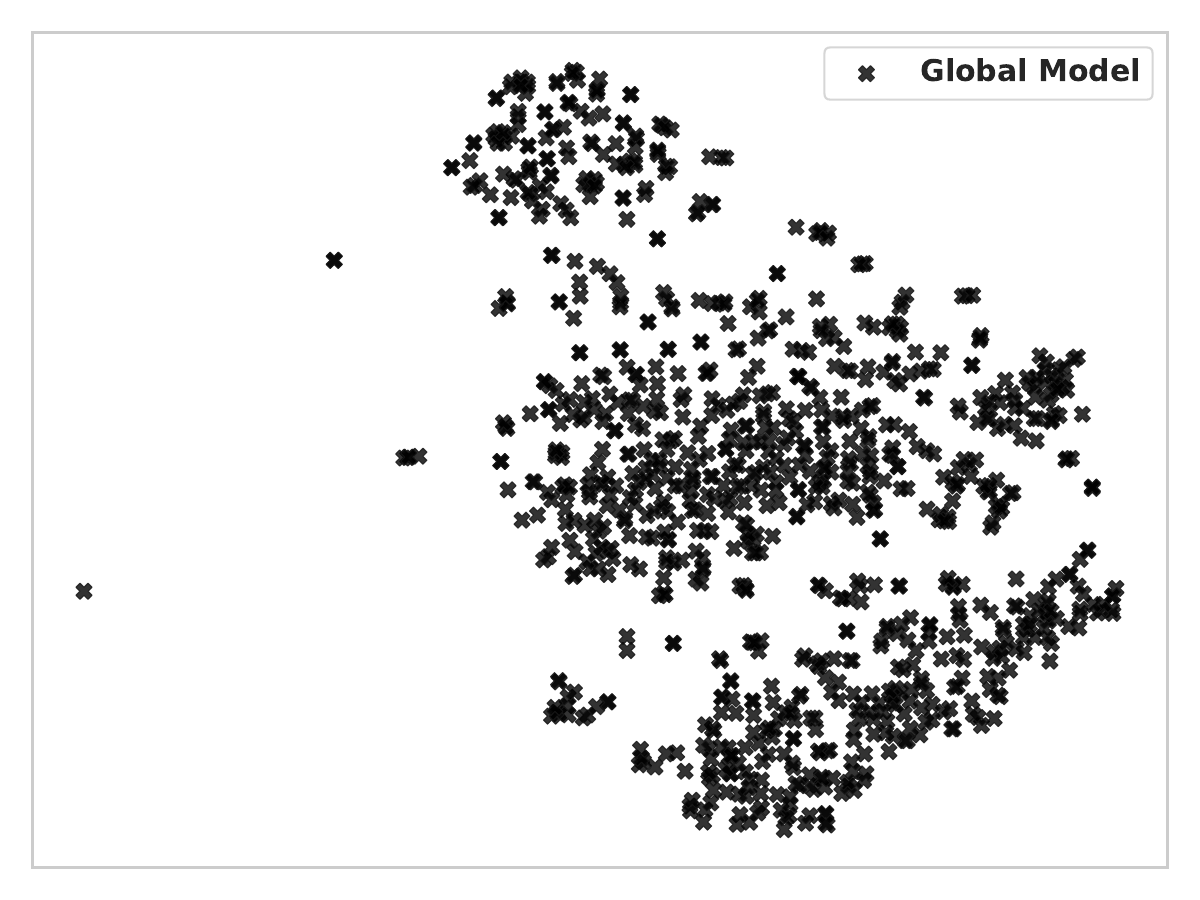}
    \textbf{(b)} After training
  \end{minipage}
  \caption{t-SNE visualization of the TB dataset before and after training \textbf{without knowledge distillation}. Left: initial feature distribution shows significant client discrepancy (non-IID). Right: after one communication round, features remain scattered, indicating \textbf{insufficient convergence} due to local distribution divergence.}
  \label{fig:tsne1}
\end{figure}

In OSFL with generative models, researchers have also adopted knowledge distillation (KD) \citep{hinton2015distilling} to facilitate global model convergence under non-IID data, as relying solely on a single round of model aggregation can introduce bias, making convergence difficult (see Figure~\ref{fig:tsne1}). In previous works \citep{zhang2022dense,kang2023one,yang2023one}, during the KD process, the global model (student model) learns not only from synthetic data but also from the output logits of the client models (teacher models). The logits from teacher models can capture diverse client data distributions, thus aiding aggregation under non-IID conditions. However, medical images exhibit greater feature differences and more complex decision boundaries compared to natural images \citep{liu2025pcrfed,kang2023one}. Moreover, potential distribution shifts in synthetic data may corrupt logit information from client models, thereby impairing aggregation performance. Thus, KD aggregation methods that rely solely on teacher logits are insufficient to ensure robust global model performance.

To address the challenges of generative model-based OSFL for medical image classification, we propose a modified framework incorporating a new Feature-Guided Rectified Flow Model (FG-RF) and a Dual-Layer Knowledge Distillation (DLKD) method. Specifically, in the FG-RF model, a feature extractor is first trained to extract features from the original client images. The Rectified Flow Model (RFM) \citep{liu2022flow} then learns the velocity field at the feature level, achieving better privacy protection with shorter training and inference times compared to DDPM. To further enhance global model aggregation, the DLKD is designed to enable the global model (student model) to not only receive the output logits of client models (teacher models) but also to align its intermediate-layer features with those of the client models. This allows the student model to capture more knowledge like feature representations, which cannot be fully transferred through final logits, helping to solve problems of feature difference in non-IID medical images and enhance the generalization ability of the global model. Moreover, the DLKD strategy helps mitigate performance degradation caused by the distribution gap between synthetic and original features.

Our contributions are summarized as follows:
\begin{itemize}[topsep = 5 pt, itemsep=2pt, parsep=0pt]
    \item We design a new OSFL framework specialized for medical image classification, incorporating diffusion model and knowledge distillation (KD). This approach uses KD to achieve single-round aggregation with synthetic data matching client distributions. It avoids multiple communication rounds in traditional FL by requiring clients to upload only their trained models.
    \item We design a new Feature-Guided Rectified Flow Model (FG-RF) to accelerate generative model speed while enhancing privacy protection in federated learning scenarios.
    \item We propose a new Dual-Layer Knowledge Distillation (DLKD) method to improve generalization under non-IID conditions and mitigate the mismatch between synthetic and original data.
    \item We provide theoretical analysis on the privacy guarantees of FG-RF and the performance benefits of DLKD.
    \item We validate our new framework and method on three medical image datasets with non-IID distributions, demonstrating its effectiveness in both accuracy and privacy preservation.
\end{itemize}

\begin{figure*}[t]
\centering
\includegraphics[width=0.9\textwidth]{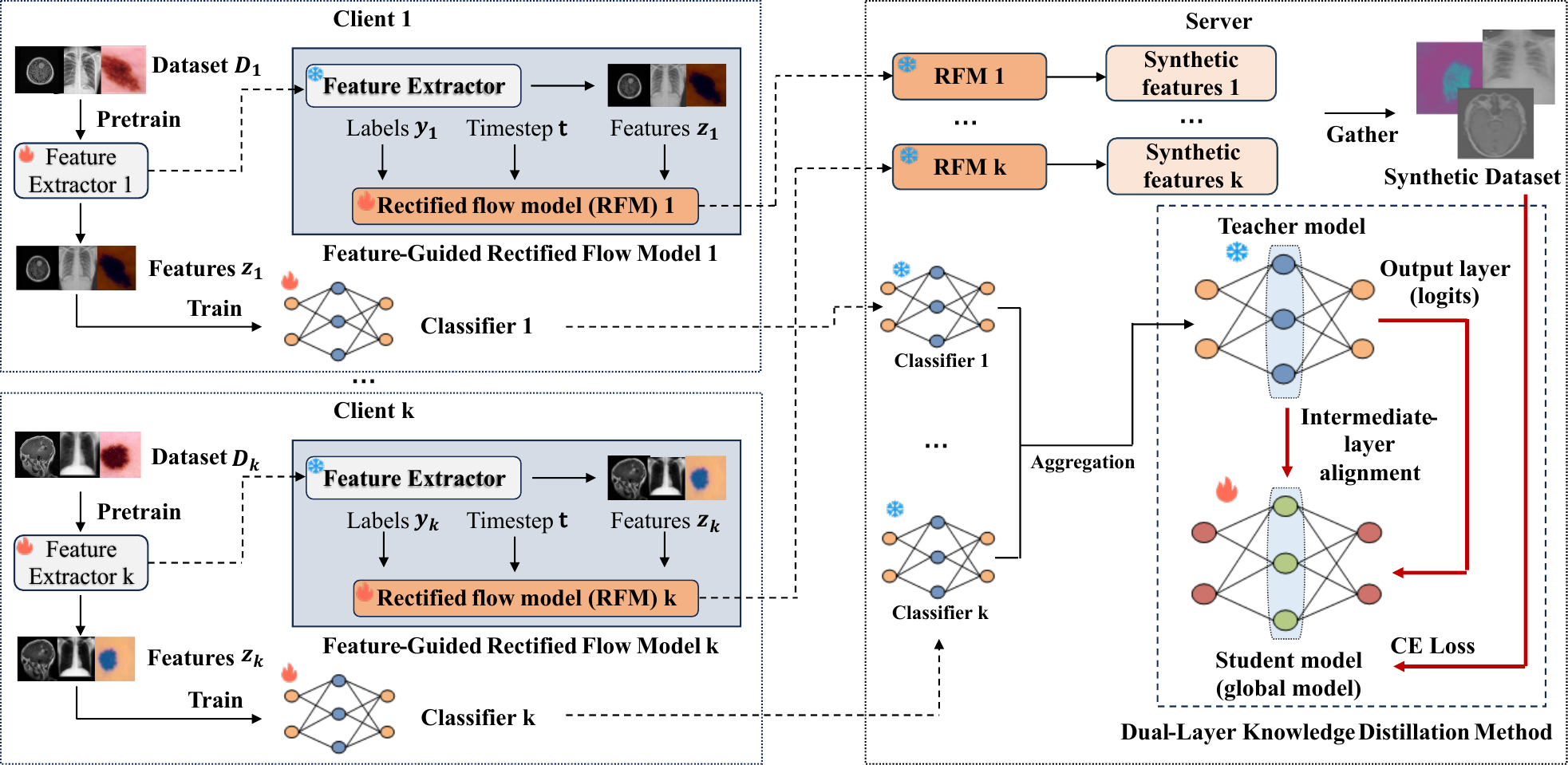}
\caption{The overall framework of our proposed method.}
\label{fig:block_diagram_final}
\end{figure*}


\section{Related works}

OSFL aims to achieve effective aggregation with minimal communication and is typically classified into three types: auxiliary dataset-based, data transmission-based, and generative model-based approaches. Early methods use server-side auxiliary datasets for aggregation \citep{guha2019one,li2020practical}, but finding matched distributions is challenging, especially for medical data, and may raise privacy concerns. To address this, some works transmit processed data instead, such as distributions \citep{kasturi2020fusion,beitollahi2024parametric}, intermediate features \citep{deng2023enhancing}, or distilled samples \citep{zhou2020distilled}, though privacy risks remain. Generative model-based methods avoid raw data transfer by sending models or prompts. For instance, \citet{heinbaugh2023data} send a VAE decoder; OSGAN \citep{kasturi2023osgan} uses client-trained GANs; and DENSE \citep{zhang2022dense}, FedISCA \citep{kang2023one}, Co-Boosting \citep{dai2024enhancing} adopt adversarial training. Yet, VAEs and GANs often generate low-quality data. To improve data quality, recent methods leverage pretrained LDMs. \citet{yang2023one} guide LDMs using client classifiers; \citet{zhang2023federated} and \citet{yang2024feddeo} transmit prompts or text descriptions; and \citet{chen2024fedbip} explore instance-level and concept-level personalization. Nonetheless, these approaches rely on pretrained LDMs for natural images, which are poorly aligned with medical image distributions. For medical imaging tasks, existing OSFL methods such as \citep{luo2024mpcpa} and \citep{zhang2025non} utilize pixel-level DDPMs for classification and segmentation. However, generating pixel-level data with diffusion models risks memorizing client-specific content \citep{carlini2023extracting}. Our FG-RF component addresses these issues by generating feature-level synthetic data tailored to medical FL settings. It enhances efficiency while preserving privacy in both training and inference.


\section{Methods}

Our proposed framework for One-Shot Federated Learning (OSFL) in medical imaging classification is mainly composed of two components: the Feature-Guided Rectified Flow Model (FG-RF) and Dual-Layer Knowledge Distillation (DLDK) method. In FG-RF, each client first trains a feature extractor to obtain features for subsequent Rectified Flow Model (RFM) training due to privacy concerns. The RFM is then trained based on the extracted features and subsequently transmitted to the server. Meanwhile, DLKD enables the global model to learn not only from the output logits of client models but also from dynamically aligned intermediate-layer features, enhancing generalization and mitigating overfitting.

This section introduces the problem formulation, including key notations and optimization objectives, followed by detailed descriptions of FG-RF and DLKD. Finally, we present a theoretical analysis demonstrating that FG-RF achieves lower privacy leakage risks compared to pixel-level RFMs, and that DLKD effectively reduces the generalization error upper bound when training on synthetic data. The overall framework is depicted in Figure~\ref{fig:block_diagram_final}, and the corresponding algorithm is outlined in Algorithm~\ref{alg:proposed_method}.

\subsection{Problem formulation}
 Assuming that we have $\mathrm{K}$ clients with their private dataset $D_K=\{(x_i,y_i)\}_{i=1}^{N^k},\ k=1, \ldots, K$, where $x_i$ represents the image, $y_i$ represents the class label, and $N^k$ is the number of images for the $k$-th client. In the FG-RF component, each client trains a feature extractor $f_{\theta_k}$ to obtain image features, which are then used to train a local Rectified Flow Model (RFM) $g_{\phi_{k}}$ and a classifier $c_{\gamma_{k}}$. The goal is to use the RFMs on the server side to synthesize a global set of image features $\mathcal{D}_{\mathrm{syn}}$, and perform DLKD training to obtain a global model $s_{\psi}$ through a single round of communication. The final optimization objective is as follows:
%
\begin{equation}
\min_{\psi}\, \mathbb{E}_{(x, y) \sim \mathcal{D}_{\mathrm{syn}}} \left[ 
\mathcal{L}_{\mathrm{CE}}(s_{\psi}(x), y) + 
\lambda\, \mathcal{L}_{\mathrm{dis}}(s_{\psi}, \bar{c}_{\gamma}) 
\right]
\end{equation}
where $\mathcal{D}_{\mathrm{syn}}$ is composed of synthetic features generated from all clients' RFMs, $\bar{c}_{\gamma}$ is the aggregated teacher model averaged from all client classifiers $c_{\gamma_{k}}$, $\mathcal{L}_{\mathrm{CE}}$ is the cross-entropy loss used to train the student model for classification, and $\mathcal{L}_{\mathrm{dis}}$ is the knowledge distillation loss.

\subsection{Feature-Guided Rectified Flow Model}

\begin{figure}[h]
\centering
\includegraphics[width=0.95\linewidth]{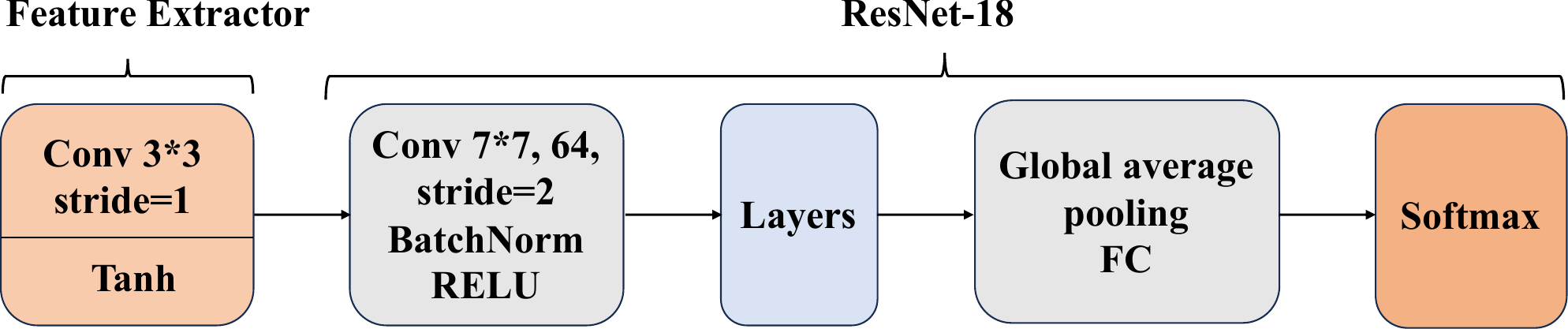}
\caption{Feature extractor training diagram.}
\label{fig:feature_extractor}
\end{figure}

\textbf{Feature Extractor Design.} The Feature-Guided Rectified Flow Model (FG-RF) is trained separately on each client. The core module of FG-RF is the feature extractor. First, each client trains a feature extractor $f_{\theta_k}$ to obtain features that guide the subsequent RFM training. To train the feature extractor, it is optimized via a classifier. In our experiments, we choose ResNet-18 \citep{he2016deep} as the classifier. The training process is illustrated in Figure~\ref{fig:feature_extractor}. Specifically, the feature extractor is directly attached to the classifier, and the training objective is to optimize the feature extractor by minimizing the cross-entropy loss $\mathcal{L}_{\mathrm{CE}}$ :
%
\begin{equation}
\mathcal{L}_{\mathrm{CE}} = -\sum y_i \log p_i
\end{equation}

The feature extractor consists of two components: a convolutional layer and the tangent activation function (tanh). The convolutional layer is chosen because it not only effectively extracts image features but also irreversibly perturbs the original image, thereby preventing excessive information leakage. In our experiments, the selected convolutional layer typically preserves the number of channels, ensuring optimal compatibility with the input requirements of the subsequent RFM network training. The design choice of the tanh(x) activation function is motivated by the need to match the input format of RFM, as its output range of $[-1, 1]$ aligns with the image data range used in diffusion models. Moreover, tanh(x) compresses the original image features, further enhancing privacy protection. The combination of the convolutional layer and the tanh(x) activation function not only enhances privacy protection but also ensures a lightweight and simple implementation, making it easy to integrate into any classification network training pipeline. 

\textbf{FG-RF Training.} After obtaining the features $z_k$ extracted from the original images using the feature extractor, each client trains its respective RFM. Unlike DDPM, the RFM models a smooth and deterministic trajectory that directly maps the randomly initialized features $z_{k,T}$ to the target features $z_{k,0}$ via an ordinary differential equation (ODE):
\begin{equation}\label{eq:my_equation7}
\frac{dz_k}{dt}=g_{\phi_k}(z_{k,t},t)
\end{equation}
here, $\frac{dz_k}{dt}$ denotes the time derivative of $z_k$ as part of the ODE. $g_{\phi_k}$ represents the learned velocity field that drives the transformation from the initial noise $z_{k,T}$ to the target data distribution $z_{k,0}$.

To model a smooth transition between the noise distribution and the data distribution, a linear interpolation scheme is used to define intermediate feature states during training. This interpolation provides a continuous trajectory from the target data features $z_{k,0}$ to the random noise $z_{k,T}$, allowing the model to learn the velocity field along this path. The interpolated state at time $t\in[0,1]$ is defined as:
\begin{equation}\label{eq:my_equation8}
z_{k,t}=(1-t)z_{k,0}+tz_{k,T}, \quad t\in[0,1]
\end{equation}

The training objective is to minimize the deviation from the optimal transport path by enabling $g_{\phi_k}$ to directly learn the velocity field that maps the noise sample $z_{k,T} \sim \mathcal{N}(\mathbf{0}, \mathbf{I})$ to the target data $z_{k,0}$, as shown in Equation~(\ref{eq:my_equation9}):
\begin{equation}\label{eq:my_equation9}
\mathcal{L}_{\mathrm{RFM}}=\mathbb{E}_{z_{k,0},z_{k,T},t}[\|g_{\phi_k}(z_{k,t},t)-(z_{k,T}-z_{k,0})\|^2]
\end{equation}
where ${z}_{k,T}-{z}_{k,0}$ denotes the ground-truth velocity direction along the linear path. The function $g_{\phi_k}$ directly regresses this optimal velocity field.

Specifically, $g_{\phi_{k}}$ is implemented using the Denoising Transformer (DiT) \citep{peebles2023scalable} network. The self-attention mechanism in Transformers captures dependencies between any two positions in the input, making it more effective at modeling global relationships. This ability to incorporate global context enables the model to coordinate content generation across the entire image more effectively, in contrast to the local receptive fields of convolution operations in U-Net. The training diagram is shown in Figure~\ref{fig:diffusion}:

\begin{figure}[h]
\centering
\includegraphics[width=0.95\linewidth]{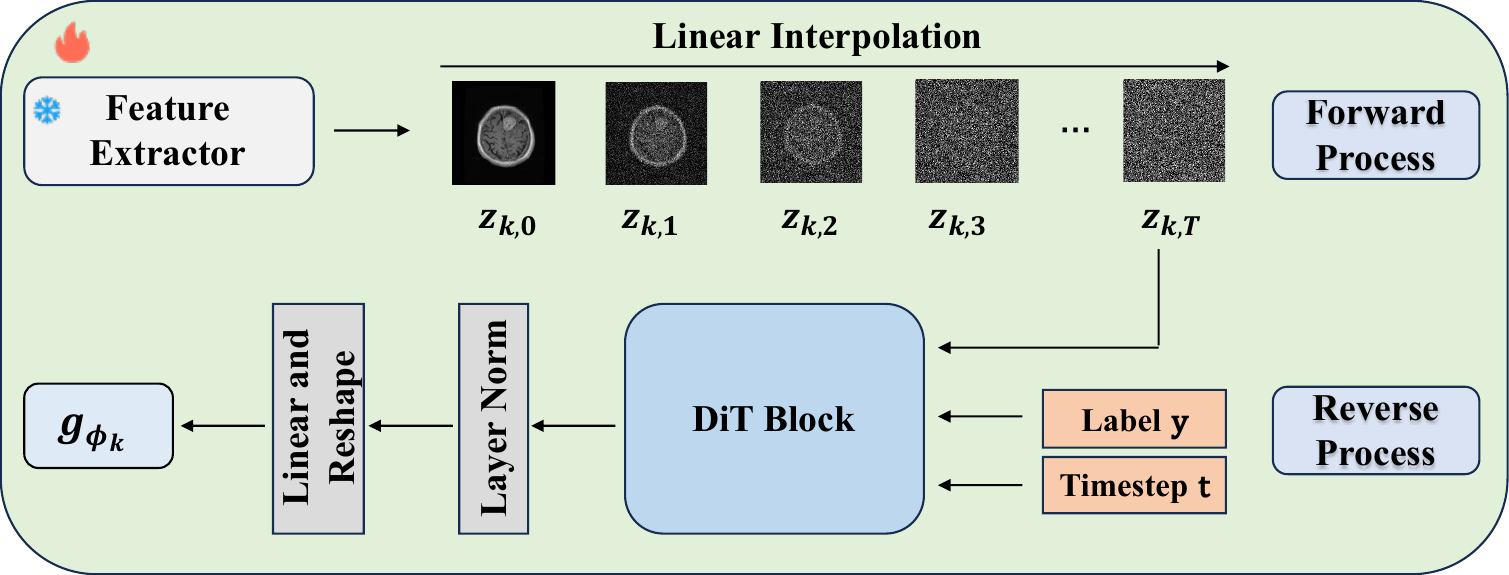}
\caption{The Rectified Flow Model training diagram.}
\label{fig:diffusion}
\end{figure}

\textbf{Features Sampling.} After completing the FG-RF training, each client will send its trained RFM model $g_{\phi_{k}}$ to the server. The server then uses the $g_{\phi_{k}}$ trained by each client to infer the image features $\hat{z}_{k}$. The sampling process is shown in Equation~(\ref{eq:my_equation6}):
\begin{equation}\label{eq:my_equation6}
\hat{z}_k=z_{k,T}+\int_T^0-g_{\phi_k}(z_k,t)dt
\end{equation}

After getting the synthetic features from each client model, they are combined into a new feature dataset $\mathcal{D}_{\mathrm{syn}}$ for subsequent server-side training.
%
\begin{equation}
\mathcal{D}_{\mathrm{syn}} = \bigcup_{k=1}^K \left\{ \hat{z}_k, y_k \right\}
\end{equation}
This dataset is synthesized based on the feature data from each client, representing the characteristics of all client data.

\subsection{Dual-Layer Knowledge Distillation}

\begin{figure}[h]
  \centering
  \begin{minipage}[b]{0.235\textwidth}
    \centering
    \includegraphics[width=\linewidth]{tsne_teacher_only3.pdf}
    \textbf{(a)} Before training
  \end{minipage}
  \hspace{0.05mm}
  \begin{minipage}[b]{0.235\textwidth}
    \centering
    \includegraphics[width=\linewidth]{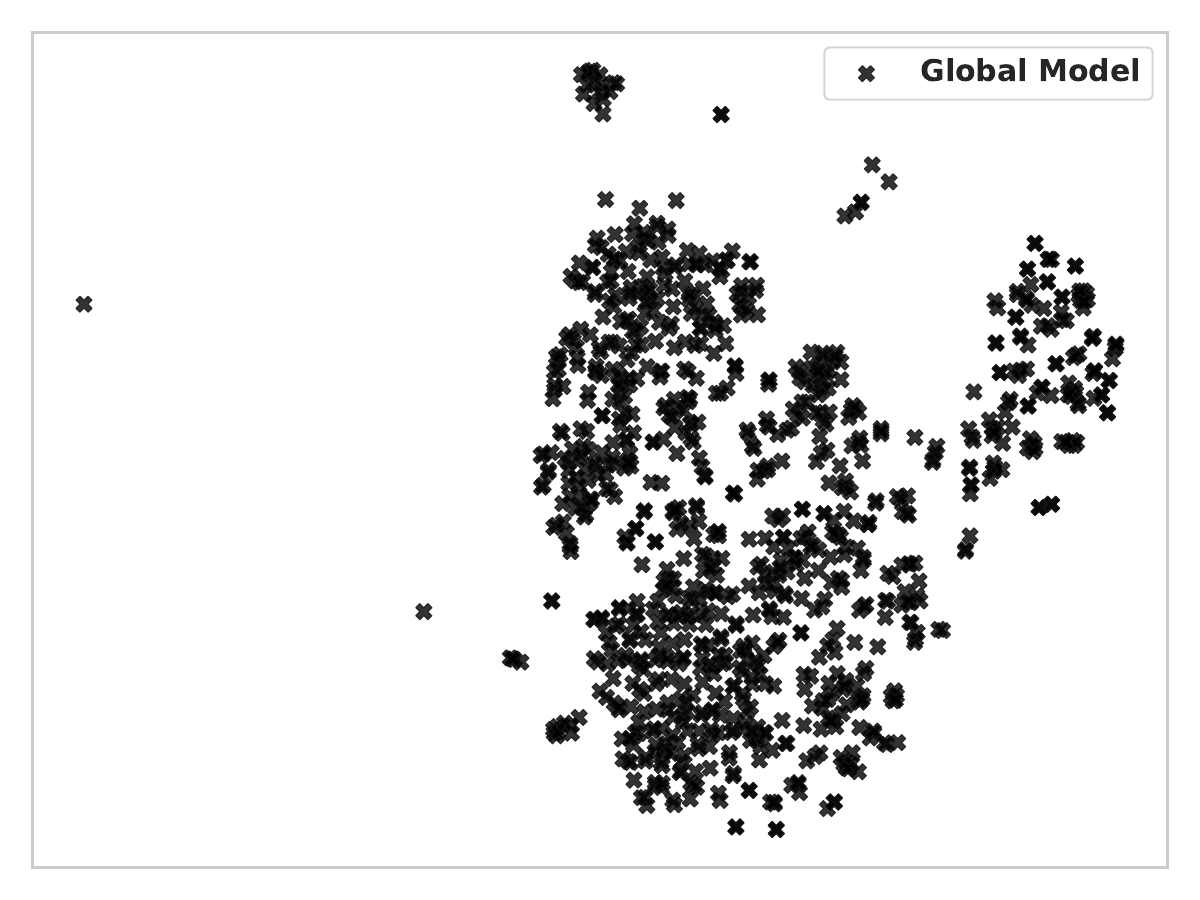}
    \textbf{(b)} After training
  \end{minipage}
  \caption{t-SNE visualization of the TB dataset before and after training \textbf{with DLKD}. }
  \label{fig:tsne2}
\end{figure}

To train the server-side student model, we employ Dual-Layer Knowledge Distillation (DLKD). Unlike prior works \citep{zhang2022dense,kang2023one,yang2023one} that only distill the output logits from client models, DLKD additionally aligns intermediate-layer features between client and student models. Specifically, the intermediate layers of the teacher model contain hierarchical feature representations, which cannot be fully transferred through the final output alone. Aligning intermediate layers allows the student model to learn better feature representations, improving convergence speed and generalization ability. Figure~\ref{fig:tsne2} shows the t-SNE visualization of the TB dataset before and after DLKD training. As shown in Figure~\ref{fig:tsne2}, the feature distributions are almost fully merged after DLKD training, implying that DLKD significantly reduces feature distribution discrepancies across clients.

Each client first trains a teacher classifier $c_{\gamma_k}$ on its real data. The input to this classifier is the features extracted by the feature extractor. This approach ensures that the extracted features are consistent with the synthetic features later used for knowledge distillation (KD) in the student model, resulting in high-quality teacher model logits. Meanwhile, it avoids directly using real data, thereby providing a certain level of privacy protection. The trained classifiers are then uploaded to the server and averaged to form an aggregation teacher model:
%
\begin{equation}
\bar{c}_\gamma = \frac{1}{K} \sum_{k=1}^K c_{\gamma_k}
\end{equation}

The student model $s_\psi$ minimizes the Kullback-Leibler (KL) divergence between its output and the teacher’s logits:
%
\begin{equation}
\mathcal{L}_{\mathrm{KL}} = \mathbb{E}_{\hat{z} \sim \mathcal{D}_{\mathrm{syn}}} \left[ D_{\mathrm{KL}}\left( \sigma\left( \frac{\bar{c}_{\gamma}(\hat{z})}{T} \right) \,\big\|\, \sigma\left( \frac{s_{\psi}(\hat{z})}{T} \right) \right) \right]
\end{equation}
%
here, $\sigma(x)$ is the softmax function, $\bar{c}_\gamma(\hat{z})$ and $s_\psi(\hat{z})$ are the output logits of the global teacher and student models for the synthetic features $\hat{z}$, respectively, and $T$ denotes the temperature parameter controlling softmax smoothness.

For feature alignment, we apply feature regression on selected intermediate layers:
%
\begin{equation}
\mathcal{L}_{\mathrm{feat}} = \mathbb{E}_{\hat{z} \sim \mathcal{D}_{\mathrm{syn}}} \left\lVert s_\psi^{(l)}(\hat{z}) - \bar{c}_\gamma^{(l)}(\hat{z}) \right\rVert_2^2
\end{equation}
%
here, $l$ represents the selected intermediate layers. In our experiments, we use ResNet-18 as an example, and $s_\psi^{(l)}(\hat{z})$ and $\bar{c}_{\gamma}^{(l)}(\hat{z})$ refer to the third-layer features of the student model $s_{\psi}$ and the teacher model $\bar{c}_{\gamma}$, respectively. We directly minimize the $L_{2}$ loss between the selected intermediate-layer features of the student and teacher models.

\begin{algorithm}[t]
\caption{Training Process of Our Proposed Method}
\label{alg:proposed_method}
\begin{algorithmic}[1]
\Require $K$ clients with local datasets $\{\mathcal{D}_k\}$, learning rates $\eta_{\mathrm{client}}$, $\eta_{\mathrm{server}}$, total epochs $T$
\Ensure Global student model $s_\psi$

\State \textbf{Initialize:} 
Each client initializes feature extractor $f_{\theta_k}$, classifier $c_{\gamma_k}$, and Rectified Flow Model $g_{\phi_k}$. 
Server initializes student model $s_\psi$.

\Statex
\State \textbf{Step 1: Client Training (parallel)}
\For{each client $k = 1$ to $K$}
    \For{each $(x_i, y_i) \in \mathcal{D}_k$}
        \State $z_i \gets f_{\theta_k}(x_i)$
        \State $\gamma_k \gets \gamma_k - \eta_{\mathrm{client}} \nabla_{\gamma_k} \mathrm{CE}(c_{\gamma_k}(z_i), y_i)$
        \State $\phi_k \gets \phi_k - \eta_{\mathrm{client}} \nabla_{\phi_k} \|g_{\phi_k}(z_i, t) - (z_{i,T} - z_{i,0})\|^2$
    \EndFor
    \State Upload $\gamma_k$, $g_{\phi_k}$ to server
\EndFor

\Statex
\State \textbf{Step 2: Server Data Inference and Teacher Model Aggregation}
\For{each client $k = 1$ to $K$}
    \State $\hat{z}_k = z_{k,T} + \int_T^0 -g_{\phi_k}(z_k, t)\, dt$
\EndFor
\State $\mathcal{D}_{\mathrm{syn}} \gets \bigcup_{k=1}^K \{(\hat{z}_k, y_k)\}$
\State $\bar{c}_\gamma \gets \frac{1}{K} \sum_{k=1}^K c_{\gamma_k}$

\Statex
\State \textbf{Step 3: Global Model Training}
\For{epoch $i = 1$ to $T$}
    \For{each mini-batch $b \sim \mathcal{D}_{\mathrm{syn}}$}
        \State $\psi \gets \psi - \eta_{\mathrm{server}} \nabla_\psi \Big( 
        (1 - \alpha) \cdot \mathrm{CE}(s_\psi(b), y)$
        \Statex \hspace{1.5em} $+ \alpha \cdot \mathrm{KL}(\sigma(\bar{c}_\gamma(b)/T) \| \sigma(s_\psi(b)/T))$
        \Statex \hspace{1.5em} $+ \beta \cdot \left\lVert s_\psi^{(l)}(b) - \bar{c}_\gamma^{(l)}(b) \right\rVert_2^2$ 
    \EndFor
\EndFor

\State Server sends $s_\psi$ back to clients
\end{algorithmic}
\end{algorithm}

Combining this with the server-side synthetic feature dataset $\mathcal{D}_{\mathrm{syn}}$, the overall loss function for training DLKD on the server is:
%
\begin{align}
\mathcal{L}_{\mathrm{total}} = \mathbb{E}_{\hat{z} \sim \mathcal{D}_{\mathrm{syn}},\, y \sim \mathcal{D}_{\mathrm{syn}}} \big[\,
& (1 - \alpha) \, \mathcal{L}_{\mathrm{CE}}(s_\psi(\hat{z}), y) 
+ \alpha \, \mathcal{L}_{\mathrm{KL}}(s_\psi(\hat{z}), \nonumber \\ &  \bar{c}_\gamma(\hat{z})) + \beta \, \mathcal{L}_{\mathrm{feat}}(s_\psi, \bar{c}_\gamma) \big]
\end{align}
here, $\mathcal{L}_{\mathrm{CE}}$ represents the cross-entropy loss, which is used to learn the hard labels $y$. The parameters $\alpha$ and $\beta$ correspond to the weights for the KL divergence loss and the intermediate-layer feature alignment loss, respectively. These hyperparameters are used to balance the influence of different loss terms.

\subsection{Theoretical analysis}
\textbf{The FG-RF privacy analysis.} The problem of quantifying privacy leakage has been extensively studied in information theory. Existing literature has proposed formal privacy metrics with strong theoretical guarantees by using mutual information (MI) \citep{bloch2021overview}. Our analysis adopts Shannon mutual information \citep{shannon1948mathematical} to measure the information leakage between original data and synthetic features.

We define $I(x;\hat{z})$ to measure the amount of shared information between the original data ${x}$ and the generated feature $\hat{z}$. A lower value of $I(x;\hat{z})$ implies that less information about ${x}$ is retained in $\hat{z}$, thereby reducing the possibility of reconstructing ${x}$ from $\hat{z}$ and suggesting a lower risk of privacy leakage. This is supported by the Nonlinear Information Bottleneck framework \citep{kolchinsky2019nonlinear}, which demonstrates that minimizing MI between the input and the latent representation effectively constrains the information that can be used to infer the original input.

Given that the original data ${x}$ is transformed into a feature ${z}$ through an irreversible and lossy feature extractor $f_{\theta}$, and $\hat{z}$ is the feature generated by FG-RF, the transition $x\to z\to\hat{z}$ depends only on the previous state and thus forms a Markov chain. According to the Data Processing Inequality (DPI) \citep{shannon1948mathematical}, which states that for any Markov chain $x\to z\to\hat{z}$, the mutual information cannot increase along the chain, we have: 
\begin{equation}\label{eq:my_equation16}
I(x;\hat{z})\leq I(x;z)
\end{equation}
Thus, the MI between the original data ${x}$ and the generated feature $\hat{z}$ cannot exceed the MI between ${x}$ and the intermediate feature ${z}$.

For the pixel-level image generation, the MI is:
\begin{equation}
I(x;\hat{x})=H(x)-H(x\mid\hat{x})
\end{equation}

Since the training objective of the RFM model is to fully learn the image generation distribution at the pixel level and minimize the error between the generated image $\hat{x}$ and the original image ${x}$, $\hat{x}$ can reconstruct ${x}$ with high fidelity. Therefore, the conditional entropy $H(x\mid\hat{x})$ is very small and can be approximated as $H(x\mid\hat{x})\approx0$. Thus, we have:
\begin{equation}\label{eq:my_equation18}
I(x;\hat{x})\approx H(x)
\end{equation}

However, for feature-level image generation, the transformation $x\to z$ inevitably leads to information loss due to the lossy nature of the feature extractor $f_{\theta}$, i.e., $H(x\mid z)>0$. According to the mutual information formula:
\begin{equation}
I(x;z)=H(x)-H(x\mid{z})<H(x)\approx I(x;\hat{x})
\end{equation}

Thus, combining with the DPI result, we conclude:
\begin{equation}
I(x;\hat{z})\leq I(x;z)<I(x;\hat{x})
\end{equation}
This indicates that the mutual information between ${x}$ and the generated feature $\hat{z}$ is strictly less than that between ${x}$ and $\hat{x}$ in pixel-level generation, suggesting a lower risk of privacy leakage in feature-level generation.	

\textbf{The analysis of the upper bound of DLKD error.} Since our OSFL framework employs synthetic data for training the final global model, we theoretically derive the upper bound of the generalization error of the global model $s_{\psi}$ on the real dataset. This analysis demonstrates that our proposed DLKD can effectively reduce the error introduced by generated data. 

Let the expected error of the global model on the real distribution be denoted as:
\begin{equation}
\varepsilon_{\mathrm{real}}(s_{\psi})=\mathbb{E}_{x\sim\mathcal{D}_{\mathrm{real}}}[\ell(s_{\psi}(x),y)]
\end{equation}
Since this quantity is generally unknown, we introduce the ensemble of client-side teacher model $\bar{c}_{\gamma}$ as an intermediate reference. By applying the triangle inequality, we obtain:
\begin{equation}
\varepsilon_{\mathrm{real}}(s_{\psi})\leq\varepsilon_{\mathrm{real}}(\bar{c}_{\gamma})+|\varepsilon_{\mathrm{real}}(s_{\psi})-\varepsilon_{\mathrm{real}}(\bar{c}_{\gamma})|
\end{equation}

Further decomposing the second term on the right-hand side:
%
\begin{align}
\varepsilon_{\mathrm{real}}(s_{\psi}) \leq\ 
& \varepsilon_{\mathrm{real}}(\bar{c}_{\gamma}) 
+ |\varepsilon_{\mathrm{real}}(s_{\psi}) - \varepsilon_{\mathrm{gen}}(s_{\psi})| \notag \\
& + |\varepsilon_{\mathrm{gen}}(s_{\psi}) - \varepsilon_{\mathrm{real}}(\bar{c}_{\gamma})|
\end{align}

To bound the first error difference term, we invoke Ben-David’s domain adaptation theory \citep{ben2010theory}, which provides the following inequality for the generalization gap of the same model under two distributions:
\begin{equation}
|\varepsilon_{\mathrm{real}}(h)-\varepsilon_{\mathrm{gen}}(h)|\leq\frac{1}{2}d_{\mathcal{H}}(\mathcal{D}_{\mathrm{real}},\mathcal{D}_{\mathrm{syn}})+\lambda
\end{equation}
here, $d_{\mathcal{H}}(\cdot,\cdot)$ denotes the $\mathcal{H}$-divergence, which quantifies the difference between the real and generated data distributions, and ${\lambda}$ represents the minimum combined error of the optimal hypothesis over both domains. Substituting this into our earlier inequality for $s_{\psi}$, we obtain:
%
\begin{align}
\varepsilon_{\mathrm{real}}(s_{\psi}) \leq\ 
& \varepsilon_{\mathrm{real}}(\bar{c}_{\gamma}) 
+ \frac{1}{2} d_{\mathcal{H}}(\mathcal{D}_{\mathrm{real}}, \mathcal{D}_{\mathrm{syn}}) 
+ \lambda \notag \\
& + |\varepsilon_{\mathrm{gen}}(s_{\psi}) - \varepsilon_{\mathrm{real}}(\bar{c}_{\gamma})|
\end{align}

This inequality reveals two key factors that influence the generalization error of the global model:
\begin{itemize}[topsep = 5 pt, itemsep=2pt, parsep=0pt]
    \item Distributional Gap: $\frac{1}{2}d_{\mathcal{H}}(\mathcal{D}_{\mathrm{real}},\mathcal{D}_{\mathrm{syn}})$ emphasizes the importance of minimizing the domain divergence between real and generated data. Smaller divergence leads to better generalization.
    \item Distillation Consistency: $|\varepsilon_{\mathrm{gen}}(s_{\psi})-\varepsilon_{\mathrm{real}}(\bar{c}_{\gamma})|$ captures the discrepancy between the global student model and the teacher ensemble under different data distributions.
\end{itemize}

Our DLKD reduces both terms. Unlike traditional distillation that only matches output logits, DLKD additionally aligns intermediate-layer features, enhancing both feature extraction and classification robustness. Feature alignment also maps generated features closer to the real data subspace, effectively reducing $d_{\mathcal{H}}$. As a result, both the distillation error and distributional gap are minimized, leading to better generalization on real-world data. 


\section{Experiment and analysis}

We conduct a series of experimental validations for our proposed method, including performance comparisons with baseline methods, efficiency comparisons between DDPM and Feature-Guided Rectified Flow Model (FG-RF), ablation studies, and privacy analysis of the FG-RF.

\subsection{Dataset preparation}

We select three different medical imaging datasets for our experiments: \textbf{1) the Chest X-Ray Tuberculosis dataset \citep{luo2024mpcpa}:} It is a real-world federated dataset collected from three different regions: China, India, and Montgomery County (USA), containing 727, 155, and 138 images. Each region includes two categories: Normal and Tuberculosis. \textbf{2) the Brain Tumor Classification dataset \citep{kaggleBrainTumor}:} It consists of Magnetic Resonance Imaging (MRI) scans classified into four categories: Benign Tumor, Malignant Tumor, Pituitary Tumor, and No Tumor, with a total of 3,264 images. \textbf{3) the HAM10000 dataset (Human Against Machine with 10,000 training images) \citep{tschandl2018ham10000}:} It is used for skin cancer classification across seven different classes, containing a total of 10,015 images. In our experiment, we set up three clients. The data across clients are non-IID. For the Brain Tumor Classification and HAM10000 datasets, we apply a shard-based partitioning strategy to ensure that each client has a different number and proportion of categories \citep{mcmahan2017communication,luo2022fedsld}. For the Chest X-Ray Tuberculosis dataset, each region naturally serves as a client.

\subsection{Baseline methods comparison}

We compare our method with eleven baseline methods grouped into three categories: \textbf{1) centralized training and multi-round Federated Learning (FL)}: Centralized training serves as the performance upper bound for comparison. We select FedAvg \citep{mcmahan2017communication}, FedProx \citep{li2020federated}, FedNova \citep{wang2020tackling}, and MOON \citep{li2021model} as representatives of multi-round FL. In the experiments, we set 100 communication rounds for multi-round FL. \textbf{2) One-Shot Federated Learning (OSFL) based on adversarial approaches}: In this category, we select DENSE \citep{zhang2022dense} and FedISCA \citep{kang2023one}, both of which employ adversarial training to train the synthesizer. It is worth noting that FedISCA is specifically designed for medical image classification. \textbf{3) OSFL based on diffusion models}: For this category, we compare our method with four approaches: FedCADO \citep{yang2023one} uses client-trained classifiers to guide server-side diffusion model inference. FGL \citep{zhang2023federated} transmits prompts generated by BLIP-2 to the pretrained diffusion model on the server. FedDEO \citep{yang2024feddeo} trains and sends text descriptions to the pretrained diffusion models on the server. The DDPM baseline generates feature-level images using a DDPM model, with all other operations consistent with our method, serving as a benchmark for generative model quality. We evaluate the classification accuracy of our proposed method against the baseline methods, as shown in Table~\ref{tab:accuracy}. The experimental details can be found in the supplementary material \citep{ma2025supplementary}.

\begin{table}[h]
\caption{Classification accuracy (\%) compared with baseline methods.}
\label{tab:accuracy}
\centering
\begin{tabular}{lccc}
\toprule
\textbf{Method} & \textbf{Brain Tumor} & \textbf{Chest X-Ray} & \textbf{HAM10000} \\
\midrule
\underline{Centralized} & 75.64 & 86.11 & 73.43 \\
FedAvg      & 36.95 & 54.88 & 55.08 \\
FedProx     & 36.89 & 49.88 & 68.24 \\
FedNova     & 30.21 & 50.12 & 68.26 \\
Moon        & 22.52 & 50.12 & 68.26 \\
DENSE       & 18.77 & 52.38 & 34.65 \\
FedISCA     & 36.77 & 53.54 & 21.50 \\
FedCADO     & 32.23 & 48.63 & 49.03 \\
FGL         & 26.66 & 48.34 & 59.22 \\
FedDEO      & 19.04 & 53.57 & 4.15  \\
DDPM        & 26.64 & 66.63 & 50.41  \\
\textbf{Ours (FG-RF)} & \textbf{60.15} & \textbf{86.32} & \textbf{70.01} \\
\bottomrule
\end{tabular}
\end{table}

Our method outperforms both multi-round FL and other OSFL methods. On the Brain Tumor Classification, Chest X-Ray Tuberculosis, and HAM10000 datasets, it achieves accuracies of 60.15\%, 86.32\%, and 70.01\%, respectively. Compared to FedAvg with 100 communication rounds, our method improves accuracy by up to 31.44\% on the Chest X-Ray Tuberculosis dataset and by 14.93\% on HAM10000, which shows the smallest improvement. Similarly, our method also outperforms other multi-round FL methods. These results demonstrate that our method significantly enhances model aggregation performance under non-IID settings.

DENSE and FedISCA perform poorly on our datasets because their generator training approaches are only suitable for low-resolution images. At 224x224 resolution, a significant amount of information is lost, severely degrading the quality of model aggregation. Our method also significantly outperforms FedCADO, FGL, and FedDEO, as fine-tuning pretrained LDMs proves less effective for medical images. In contrast, our FG-RF method achieves excellent performance on medical imaging tasks. Compared to the DDPM backbone, our method improves accuracy by an average of 24.27\%, demonstrating that the Rectified Flow Model (RFM) can generate higher-quality feature-level images than DDPM.

It is noteworthy that FedISCA is the only OSFL method specifically designed for medical images, and our approach surpasses it by up to 48.51\% on the HAM10000 Dataset. FedISCA uses adversarial training to generate synthetic data on the server and applies knowledge distillation based solely on logits. In contrast, our method leverages FG-RF for data synthesis and adopts Dual-Layer Knowledge Distillation (DLKD), learning from both logits and intermediate features, leading to significantly improved performance in medical image classification.

\begin{table*}[t]
\caption{Comparison of synthetic results across different methods.}
\label{tab:images}
\centering
\begin{tabular}{lccccccccc}
\toprule
\textbf{Dataset} & \textbf{Original} & \makecell{\textbf{DENSE} \\ \textbf{(features)}} & \makecell{\textbf{FedISCA} \\ \textbf{(features)}} & \textbf{FedCADO} & \textbf{FGL} & \textbf{FedDEO} & \makecell{\textbf{DDPM} \\ \textbf{(features)}} & \makecell{\textbf{Ours} \\ \textbf{(images)}} & \makecell{\textbf{Ours} \\ \textbf{(features)}} \\
\midrule
Brain Tumor &
\includegraphics[width=0.06\linewidth]{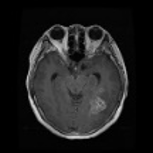} &
\includegraphics[width=0.06\linewidth]{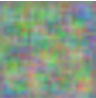} &
\includegraphics[width=0.06\linewidth]{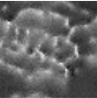} &
\includegraphics[width=0.06\linewidth]{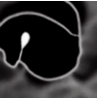} &
\includegraphics[width=0.06\linewidth]{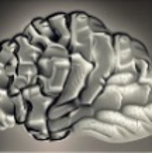} &
\includegraphics[width=0.06\linewidth]{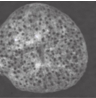} &
\includegraphics[width=0.06\linewidth]{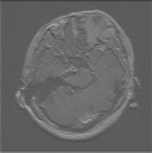} &
\includegraphics[width=0.06\linewidth]{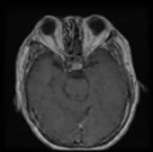} &
\includegraphics[width=0.06\linewidth]{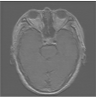} \\
\midrule
Chest X-Ray &
\includegraphics[width=0.06\linewidth]{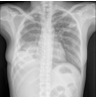} &
\includegraphics[width=0.06\linewidth]{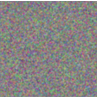} &
\includegraphics[width=0.06\linewidth]{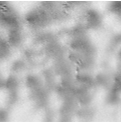} &
\includegraphics[width=0.06\linewidth]{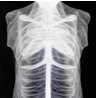} &
\includegraphics[width=0.06\linewidth]{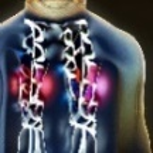} &
\includegraphics[width=0.06\linewidth]{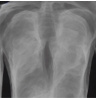} &
\includegraphics[width=0.06\linewidth]{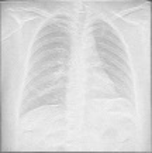} &
\includegraphics[width=0.06\linewidth]{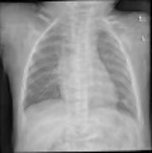} &
\includegraphics[width=0.06\linewidth]{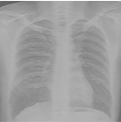} \\
\midrule
HAM10000 &
\includegraphics[width=0.06\linewidth]{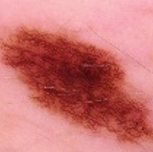} &
\includegraphics[width=0.06\linewidth]{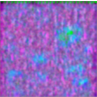} &
\includegraphics[width=0.06\linewidth]{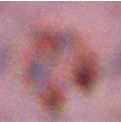} &
\includegraphics[width=0.06\linewidth]{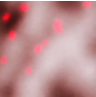} &
\includegraphics[width=0.06\linewidth]{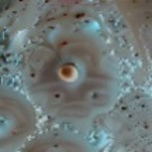} &
\includegraphics[width=0.06\linewidth]{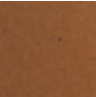} &
\includegraphics[width=0.06\linewidth]{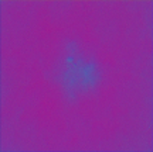} &
\includegraphics[width=0.06\linewidth]{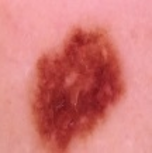} &
\includegraphics[width=0.06\linewidth]{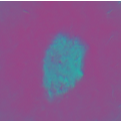} \\
\bottomrule
\end{tabular}
\end{table*}

Table~\ref{tab:images} compares the image features generated by our method with the original images and baseline methods (See supplementary material for more results \citep{ma2025supplementary}). DENSE, FedISCA, DDPM and our method generate image features, while the others synthesize pixel-level images. The results show that DENSE and FedISCA struggle to produce high-resolution features, resulting in detail loss and impaired model aggregation. FedCADO, FGL, and FedDEO generate synthetic data with large distribution shifts, also degrading aggregation performance. In contrast, our method generates high-quality features with slight blurring, effectively balancing utility and privacy.
\subsection{Efficiency Comparison: DDPM vs. FG-RF in Training and Inference}

To demonstrate the efficiency of FG-RF, we compare the training and inference times of DDPM and FG-RF, as shown in Table~\ref{tab:time_comparison}. The results show that FG-RF significantly reduces both training and inference times across all three datasets, indicating that the RFM backbone is more suitable for OSFL scenarios.


\begin{table}[h]
\caption{The comparison of training and inference time between DDPM and FG-RF.}
\label{tab:time_comparison}
\centering
\begin{tabular}{lccc}
\toprule
\textbf{Dataset} & \textbf{Model} & \makecell{\textbf{Training Time} \\ \textbf{(800 epochs)}} & \makecell{\textbf{Inference Time} \\ \textbf{(1 image)}} \\
\midrule
\multirow{2}{*}{\makecell{Brain Tumor}} 
    & DDPM   & 4.51 h  & 22.68 s \\
    & FG-RF  & 1.87 h  & 0.45 s  \\
\midrule
\multirow{2}{*}{\makecell{Chest X-Ray}} 
    & DDPM   & 1.36 h  & 29.32 s \\
    & FG-RF  & 0.58 h  & 0.69 s  \\
\midrule
\multirow{2}{*}{HAM10000} 
    & DDPM   & 12.92 h & 18.88 s \\
    & FG-RF  & 5.41 h  & 0.29 s  \\
\bottomrule
\end{tabular}
\end{table}

\subsection{Ablation Study}

To verify the contribution of each proposed component, we conduct ablation experiments on the intermediate-layer alignment and DLKD, as shown in Table~\ref{tab:ablation}. The results demonstrate that all components contribute significantly to the overall performance improvement.

\begin{table}[h]
\caption{Ablation experiment.}
\label{tab:ablation}
\centering
\begin{tabular}{lccc}
\toprule
\textbf{Method} & \makecell{\textbf{Brain Tumor}} & \makecell{\textbf{Chest X-Ray}} & \makecell{\textbf{HAM10000}} \\
\midrule
\makecell[l]{w/o DLKD and feature \\ extractor }    & 54.56 & 77.09 & 68.97 \\
w/o DLKD                          & 55.08 & 83.16 & 64.91 \\
\makecell[l]{w/o intermediate-layer \\ alignment} & 55.84 & 84.74 & 68.99 \\
\textbf{Ours}                             & \textbf{60.15} & \textbf{86.32} & \textbf{70.01} \\
\bottomrule
\end{tabular}
\end{table}

\subsection{Privacy analysis of FG-RF}

To verify that generating feature-level images reduces privacy leakage compared to pixel-level images, we follow the methodology of \citep{carlini2023extracting} and evaluate the privacy risk of diffusion models by measuring their memorization behavior. Specifically, we compute the Euclidean 2-norm distance between original and generated images, defined as $\ell_2(a,b) = \sqrt{\sum_i(a_i-b_i)^2/d}$, where $d$ is the input dimension. A training image $x$ is considered to be memorized by the diffusion model if there exists a generated image $\hat{x}$ such that $\ell(x, \hat{x}) \leq \delta$, where the threshold $\delta$ is set to 0.1 following \citep{carlini2023extracting}.

Since our method generates feature-level images rather than pixel-level images directly, we first design and train an image feature decoder to reconstruct the image from the synthetic features. We then calculate the $\ell_{2}$ distances for both feature-level and pixel-level generations. The results (see Table~\ref{tab:l2_distance}) demonstrate that both the feature-reconstructed images and the pixel-level generated images maintain $\ell_2$ distances greater than 0.1 from their corresponding original images, meeting the privacy requirements. Moreover, our method achieves larger distances compared to directly generating pixel-level images, highlighting that the proposed FG-RF method offers enhanced privacy protection.

\begin{table}[h]
\caption{$\ell_{2}$ distance comparison for diffusion model generation.}
\label{tab:l2_distance}
\centering
\begin{tabular}{llccc}
\toprule
\textbf{Methods} & \textbf{Client ID} & \textbf{Brain Tumor} & \makecell{\textbf{Chest X-Ray}} & \textbf{HAM10000} \\
\midrule
\multirow{3}{*}{\makecell{Train with \\ pictures}} 
    & Client 0 & 0.3793 & 0.4123 & 0.2787 \\
    & Client 1 & 0.3814 & 0.4049 & 0.2534 \\
    & Client 2 & 0.4979 & 0.6304 & 0.2795 \\
\midrule
\multirow{3}{*}{\makecell{Train with \\ features}} 
    & Client 0 & 0.5195 & 0.5239 & 0.3238 \\
    & Client 1 & 0.5132 & 0.5059 & 0.2559 \\
    & Client 2 & 0.5100 & 0.7734 & 0.2872 \\
\bottomrule
\end{tabular}
\end{table}

\section{Conclusions}

In this work, we propose a novel One-Shot Federated Learning (OSFL) framework for medical imaging to enhance privacy protection and training efficiency. We design the Feature-Guided Rectified Flow Model (FG-RF) to learn from feature-level images instead of pixel-level images, effectively reducing privacy risks and accelerating local training. In addition, we develop Dual-Layer Knowledge Distillation (DLKD) method to dynamically align output logits and intermediate-layer features during global model aggregation, improving generalization across heterogeneous medical data. Theoretical analysis proves that FG-RF lowers the risk of privacy leakage, while DLKD reduces the upper bound of the generalization error. Experiments on three medical imaging datasets demonstrate that our new framework and method significantly outperform traditional multi-round federated learning and existing generative model-based OSFL approaches, while providing enhanced privacy protection.

\begin{ack}
This work is supported by 2023 Shenzhen sustainable supporting funds for colleges and universities (20231121165240001), Shenzhen Science and Technology Program (JCYJ20230807120800001), Guangdong
Provincial Key Laboratory of Ultra High Definition Immersive Media Technology (2024B1212010006).
\end{ack}


\bibliography{mybibfile}

\end{document}